\documentclass{article} 
\usepackage[final]{colm2025_conference} 

\usepackage{microtype}
\usepackage{hyperref}
\usepackage{url}
\usepackage{tabularx}
\usepackage{booktabs}
\usepackage{amssymb}  
\usepackage{xcolor}   
\usepackage{pifont}
\usepackage{lineno}
\usepackage{graphicx}
\usepackage{multirow}
\usepackage{longtable}

\definecolor{darkblue}{rgb}{0, 0, 0.5}
\hypersetup{colorlinks=true, citecolor=darkblue, linkcolor=darkblue, urlcolor=darkblue}

\title{MapIQ: Benchmarking Multimodal Large Language Models for Map Question Answering}

\author{Varun Srivastava \& Fan Lei \\
School of Computing and Augmented Intelligence \\
Arizona State University \\
Tempe, AZ 85281, USA \\
\texttt{\{vsriva11,flei5\}@asu.edu}
\AND
Srija Mukhopadhyay \\
Department of Computer Science \\
International Institute of Information Technology \\
Hyderabad, India \\
\texttt{srija.mukhopadhyay@iit.ac.in}
\AND
Vivek Gupta \& Ross Maciejewski \\
School of Computing and Augmented Intelligence \\
Arizona State University \\
Tempe, AZ 85281, USA \\
\texttt{\{vgupt140,rmacieje\}@asu.edu}
}

\newcommand{\cmark}{\textcolor{green}{\ding{51}}}
\newcommand{\xmark}{\textcolor{red}{\ding{55}}}

\begin{document}

\ifcolmsubmission
\linenumbers
\fi

\maketitle

\begin{abstract}
Recent advancements in multimodal large language models (MLLMs) have driven researchers to explore how well these models read data visualizations, e.g., bar charts, scatter plots. More recently, attention has shifted to visual question answering with maps (Map-VQA). However, Map-VQA research has primarily focused on choropleth maps, which cover only a limited range of thematic categories and visual analytical tasks. To address these gaps, we introduce MapIQ, a benchmark dataset comprising 14,706 question-answer pairs across three map types—choropleth maps, cartograms, and proportional symbol maps spanning topics from six distinct themes (e.g., housing, crime). We evaluate multiple MLLMs using six visual analytical tasks, comparing their performance against one another and a human baseline. An additional experiment examining the impact of map design changes (e.g., altered color schemes, modified legend designs, and removal of map elements) provides insights into the robustness and sensitivity of MLLMs, their reliance on internal geographic knowledge, and potential avenues for improving Map-VQA performance. 
\end{abstract}

\section{Introduction}

Multimodal Large Language Models (MLLMs) can process and reason across multiple modalities, including text, images, and structured data, enabling applications in various domains, from general visual understanding to specialized fields like medical imaging~\citep{zhou2023survey, zhang2023llavar}. Recently, researchers have begun exploring how effectively these models interpret data visualizations, specifically evaluating their ability to understand and reason about scatter plots, bar graphs, line charts, etc~\citep{kafle2018dvqa, masry2022chartqa}. Building upon this interest, Map Question Answering (Map-VQA) has emerged to assess the capabilities of MLLMs in reading geospatial visualizations. 
However, current Map-VQA research predominantly focuses on choropleth maps~\citep{mukhopadhyay2024mapwise, chang2022mapqa, slocum2022thematic} and typically evaluates simple visual analytic (VA) tasks. Moreover, these simple VA tasks often overlook the broader range of analytical tasks described in Information Visualization and Geographic Information Systems literature~\citep{munzner2014visualization, amar2005low, maceachren2004maps, brewer2016designing}. Additionally, cognitive science literature indicates thematic content can influence human map-reading accuracy due to prior biases~\citep{herrmann1996cognitive}, yet it remains unclear whether MLLMs exhibit similar thematic biases.

Motivated by these limitations, we introduce MapIQ, a benchmark dataset comprising \textbf{14,706} question-answer pairs designed to gauge the map-reading capabilities of state-of-the-art MLLMs. MapIQ introduces two previously unexplored map types—cartograms (specifically hexbin maps, which use uniform shapes to reduce visual bias from varying geographic unit sizes~\citep{fan2024understanding}) and proportional symbol maps (which encode data through varying symbol sizes instead of color gradients~\citep{slocum2022thematic})—in addition to traditional choropleth maps. MapIQ incorporates six VA tasks across local (tasks involving individual states or small regions) and global spatial scales (tasks requiring synthesis across entire maps), aligning with varying analytical complexity critical in geospatial analysis~\citep{maceachren2004maps}. MapIQ also encompasses metadata sourced from six thematic categories, allowing exploration of how thematic content affects MLLM performance.

Using the MapIQ benchmark, we evaluate seven MLLMs, both closed-source and open-source, to address: Are MLLMs biased toward choropleth maps due to their prevalent use in training datasets? How does MLLM performance vary across different VA tasks? Does thematic content significantly impact map-reading accuracy? What performance gaps exist between open-source and closed-source models in Map-VQA? Additionally, we establish a human performance baseline, examining the alignment between model-generated answers and expert human readers. Finally, aligned with recent research~\citep{wu2024chartinsights, mukhopadhyay2024unraveling}, we investigate the robustness and sensitivity of MLLMs to variations in visual elements, such as map legends and color schemes.


\section{Related Work}
\label{related_work}

\textbf{VA Tasks and Chart-VQA:} VA tasks are broadly categorized into low-level and high-level tasks~\citep{brehmer2013multi}. Low-level tasks involve visual queries, such as retrieving values or comparing data points, and rely solely on visual information without requiring broader contextual knowledge. High-level tasks demand deeper engagement, such as identifying trends and patterns, and often require domain expertise and nuanced interpretation~\citep{brehmer2013multi}. Low-level VA tasks have been widely used to assess visualization literacy in humans~\citep{lee2016vlat, pandey2023mini}. Building on this, Chart-VQA studies emerged to evaluate MLLMs' ability to answer low-level analytical questions about charts~\cite{masry2022chartqa}. Recent progress in MLLMs, e.g., ChatGPT~\citep{openai_chatgpt4o_2024} and Claude~\citep{anthropic_claude_2024}, has renewed interest in this area~\citep{bendeck2024empirical, xu2024exploring}, and recent works~\cite{wu2024chartinsights,mukhopadhyay2024unraveling} assess MLLM proficiency across various low-level VA tasks, probing their robustness to design variations. Our study extends these established tasks to geospatial contexts~\citep{maceachren2004maps, slocum2022thematic} to evaluate the capabilities and robustness of MLLMs in map-reading tasks.

\textbf{Map-VQA:} Studies in Map-VQA have emerged with approaches spanning both high-level and low-level map-reading tasks. High-level studies have evaluated MLLMs' capabilities in complex spatial reasoning tasks such as pathfinding and geolocation detection~\citep{xing2025can, roberts2024charting, hochmair2024correctness}. These efforts rely on specialized maps, including remote sensing imagery and navigation maps, and require domain-specific knowledge and advanced spatial interpretation. In contrast, research on low-level VA tasks involving thematic maps—cartographic representations that use simple visual elements like colors and symbols to depict spatial distributions of specific themes~\citep{slocum2022thematic}—has been relatively limited. Pioneering work from \citet{chang2022mapqa} introduced a dataset of choropleth maps representing data from a single theme (healthcare), evaluating MLLMs on three core tasks: basic map-reading literacy, value retrieval, and identifying spatial extremes. More recently, MapWise~\citep{mukhopadhyay2024mapwise} expanded on this by benchmarking state-of-the-art MLLMs, incorporating counterfactual testing and comparing model performance with human readers, though still limited to choropleth maps. Building on these foundations, our work presents MapIQ, a comprehensive dataset featuring multiple thematic map types across diverse themes. 

\section{MapIQ Dataset}
\label{dataset}

In this section, we detail the process followed to create the MapIQ dataset, including the metadata, map selection and generation, question generation, and data validation.

\subsection{Metadata}
In our study, one of the goals was to assess whether the thematic content of maps influenced the performance of MLLMs in Map-VQA. We identified six representative themes commonly visualized using thematic maps: social, economic, health, crime, environment, and housing~\citep{dent1999cartography, slocum2022thematic}. 
After theme selection, we sourced datasets for map generation, focusing only on the USA. Data related to social, economic, and housing themes were obtained from the U.S. Census Bureau~\citep{us_census_acs}; environmental data were sourced from the Environmental Protection Agency (EPA)~\citep{epa_data} and weather.gov~\citep{weather_gov}; crime data were collected from the FBI Crime Data Explorer~\citep{fbi_crime_data_explorer}, and health-related data were gathered from the Centers for Disease Control and Prevention (CDC)~\citep{cdc_data}. Our metadata collection process resulted in 258 datasets across six distinct themes, and full dataset details are provided in Appendix~\ref{A1.1_metadata_details}.

\begin{figure}[t]
\centering	
\includegraphics[width=\linewidth]{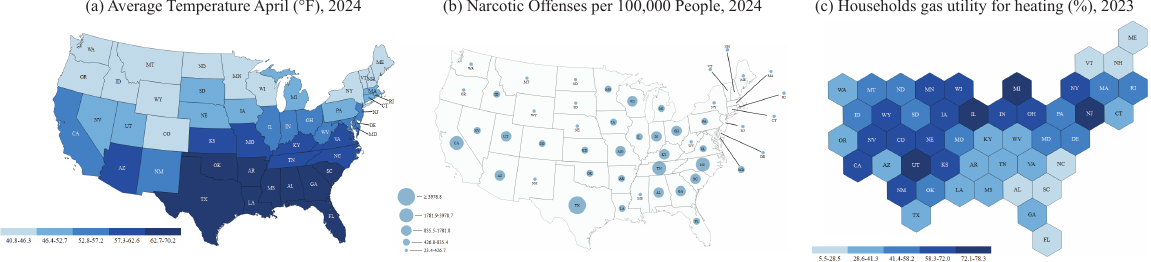}
\caption{Three baseline maps. (a) Choropleth, (b) Proportional Symbol, and (c) Cartogram}
\label{fig:map-examples}
\vspace{-5mm}
\end{figure}

\subsection{Baseline Map Design}
MapIQ explores three map types: choropleth, cartogram, and proportional symbol (Fig~\ref{fig:map-examples}).
For all map renderings, we adhered to established cartographic best practices~\citep{slocum2022thematic}. We chose discrete classification, which is the most commonly used method in thematic mapping~\citep{slocum2022thematic}, categorizing data into five classes using Fisher–Jenks classification~\citep{jenks1971error}. For the choropleth and cartogram maps, we employed a sequential blue color scheme from ColorBrewer~\citep{harrower2003colorbrewer}. For proportional symbol maps, we selected circles with a range-graded size variation~\citep{slocum2022thematic}. All maps were annotated with the official two-letter U.S. state abbreviations~\citep{stateAbbrev}, and all maps are rendered utilizing the Albers USA projection~\citep{snyder1982map} on a white background. Following best practices, data were normalized across maps, and leader lines were used for clear labeling in cases of small spatial units~\citep{slocum2022thematic}. These design considerations resulted in 258 (datasets) × 3 (map types) = 774 unique map images. We plan to open-source our map generation pipeline to facilitate the development of new Map-VQA benchmarks and enable MLLMs to conduct further research into map comprehension.

\subsection{Task Selection, Question Generation, and Ground Truth Extraction}
\label{task_question_types}

To assess how effectively MLLMs extract geospatial information at local (specific points or regions) and global (entire map) scales~\citep{maceachren2004maps}, we selected six VA tasks from established literature: Retrieve Value, Pairwise Point Comparisons, Spatial Extremes, Spatial Clusters, Determine Range, and Regional Comparisons~\citep{lee2016vlat, amar2005low, munzner2014visualization, maceachren2004maps}. For tasks requiring local-level analysis on a regional scale, we divided the 49 contiguous U.S. states into four zones: West, Midwest, Northeast, and South~\citep{uscb_econ_geo_levels}. Detailed definitions of each task type are provided in Appendix~\ref{A1.2_task_type_definitions}.
We generated questions in four formats—binary (True/False), multiple-choice (MCQ), single value, and list—aligned with prior literature~\citep{mukhopadhyay2024mapwise, lu2022learn}. Binary and MCQ formats were used for all tasks, while inherent differences among tasks required selective use of single-value (two tasks) and list formats (five tasks). Each question was carefully created using manually validated templates, resulting in 19 unique questions per map. This process yielded 774 maps × 19 questions per map = 14,706 question-answer pairs. Ground truth answers to all questions were extracted programmatically using Python. A human expert carefully reviewed the final question-answer dataset to ensure consistency and accuracy. The complete MapIQ dataset is provided in the supplementary materials, while the question templates, example questions, and additional details on the ground truth extraction process are provided in Appendices \ref{A1.3_question_templates_and_task_eligibility} and \ref{A1.4_ground_truth_extraction}, respectively. A comparative overview with existing Map-VQA benchmarks appears in Appendix \ref{A1.5_comaring_with_existing_benchmarks}.




\section{Benchmarking MLLMs with MapIQ}
\label{experiment details} 


For our experiments, we considered both closed- and open-source models. Closed-source models, benefiting from a higher parameter count, typically achieve superior VQA performance, and we investigate whether this advantage persists in the MapIQ dataset. We selected ChatGPT-4o~\citep{openai_chatgpt4o_2024}, Gemini 1.5 Pro~\citep{team2024gemini}, and Claude 3.5 Sonnet~\citep{anthropic_claude_2024}, given their impressive multimodal capabilities.
For open-source models, we focused on recently released (2024) models with robust documentation hosted on HuggingFace~\citep{huggingface_2024}. To ensure fairness and manage computational resources, we constrained our selection to models within the 7B–8B parameter range, choosing Qwen2-VL~\citep{Qwen2VL}, Molmo~\citep{deitke2024molmo}, InternVL2.5-MPO~\citep{chen2024internvl}, and Idefics3~\citep{laurençon2024building}. DeepSeek~\citep{wu2024deepseekvl2mixtureofexpertsvisionlanguagemodels} and MiniCPM-V 2.6~\citep{yao2024minicpm} were initially tested but excluded due to incoherent or missing outputs. 

\noindent \textbf{Prompting Strategy} - The prompts for benchmarking MLLMs were designed for this experiment in a zero-shot setting~\citep{radford2019language}, and each prompt was tailored according to task type and question type. Each prompt comprised two primary components. The first component included general instructions, providing a brief overview of the map-reading task that the model was expected to undertake. Additionally, this component specified instructions for formatting the model's responses according to the question type. For the Spatial Clusters task, these general instructions were further supplemented with a clear definition of spatial clusters, details about the spatial adjacency rule, and the minimum cluster size.
The second component of the prompt was designed for tasks requiring zoning information (e.g., spatial extremes, regional comparisons). For these tasks, details about geographic zones~\citep{usZoning} and their constituent states were provided first, followed by the previously described general instructions. Prompt details for all task types are available in Appendix~\ref{A2.1_prompt_details}.

\noindent \textbf{Test Dataset} - To manage computational resources, we used a representative subset of the full MapIQ dataset as our test set. We selected 35\% of the entire dataset, resulting in 5130 QA pairs. To ensure this sample was representative and balanced across the experimental variables, we employed stratified random sampling, considering map type, task type, question type, and theme. More information about the sampling process is included in Appendix~\ref{A2.2_test_set_sampling}, and the complete test dataset is provided in supplementary materials.

\noindent \textbf{Evaluation Metrics} - Given the diversity of question types in our benchmarking experiment, we tailored evaluation metrics specifically for each type. For binary questions, accuracy was selected, and scores of 100 (correct) or 0 (incorrect) were assigned and aggregated across the dataset. For MCQ-type questions, where tasks could have multiple correct options, we employed the F1 score due to its balanced treatment of precision and recall, rewarding partial correctness and penalizing false positives and negatives~\citep{van1974foundation, derczynski2016complementarity}. Similarly, list-type questions utilized F1 scores, effectively handling partial matches. Individual F1 scores were calculated per response and averaged for overall evaluation. For single-value questions, we followed the same scoring method as binary questions.

\noindent \textbf{MLLM Response Extraction and Validation} - The model responses were extracted using Python scripts on a Linux-based system equipped with an NVIDIA A100 GPU and 128 GB of RAM. We manually validated outputs to ensure quality before evaluating them with previously described metrics. During validation, some models provided contradictory responses by selecting "None of the above" (NOTA) alongside other options (e.g., "My answer is [a, b, e]," where e is NOTA), which we marked incorrect to maintain evaluation integrity. Additionally, several models frequently included unsolicited explanatory text or reasoning in their responses. Removing these extraneous details was particularly time-consuming, especially for InternVL2.5-MPO and Molmo (open-source models), and Gemini 1.5 Pro (closed-source model). The evaluated and validated datasets for each model are provided in the supplementary materials, while Appendix \ref{A2.3_mllm_response_validation} contains additional details about the MLLM response validation process.

\noindent \textbf{Human Baseline Evaluation} - Along with comparing the performance of various MLLMs, we aimed to establish how these models compare to expert human performance in the Map-VQA context. We established a human baseline by uniformly sampling around 9\% of the test set, resulting in 450 unique QA pairs. Special care was taken to ensure approximately balanced representation across all experimental variables to guarantee fairness in the human evaluation. Using this sample, two expert human map readers independently answered all 450 questions. Subsequently, a third independent expert validated their responses to ensure consistency and reliability. Humans received the same questions as those posed to the MLLMs, and identical instructions were provided across evaluations.


\section{Benchmarking MLLMs with MapIQ - Results and Discussions}
\begin{figure}[t]
\centering	
\includegraphics[width=\linewidth]{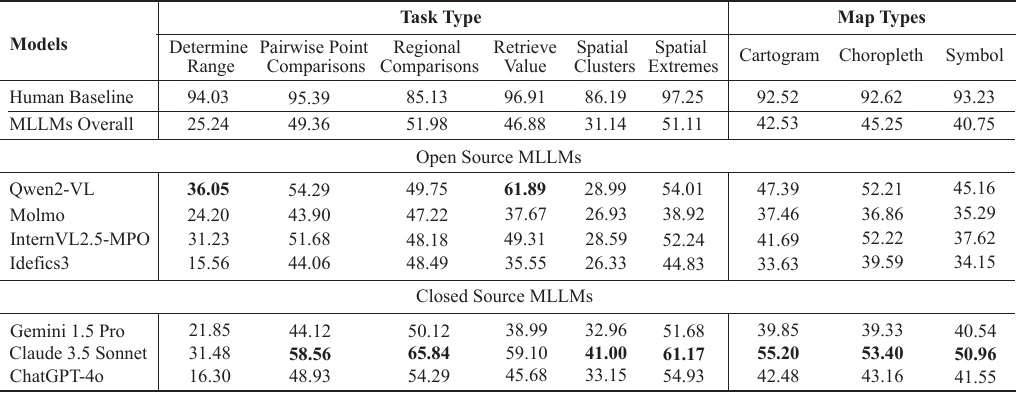}
\caption{Performance of Humans and 7 MLLMs (overall and individual) across 6 Task Types and 3 Map Types. Values represent average performance scores (in \%) calculated separately for each task and map type. Bolded values indicate the best-performing model for each experimental condition.}
\label{tab:overall-performance-task-type-question-type}
\vspace{-3mm}
\end{figure}

Detailed performance results across the MapIQ dataset are provided in Fig~\ref{tab:overall-performance-task-type-question-type} and Fig~\ref{tab:overall-performance-map-type-theme}. In this section, we discuss key findings comparing models to humans and stratifying results across task types, map types, question types, and themes.


\noindent \textbf{Comparing MLLMs Across Experimental Variables} - Overall,  Claude 3.5 Sonnet consistently outperformed other models across all four experimental variables—Task Type, Map Type, Theme, and Question Type—clearly establishing itself as the most robust MLLM evaluated in this study. It achieved top rankings in most tasks and settings, demonstrating strong adaptability to various map-reading challenges. Additionally, Qwen2-VL, an open-source model, also showed competitive results, emerging as the second-best model overall, making it a compelling open-source alternative. Detailed analysis per experimental variable is given below:

\begin{figure}[t]
\centering	
\includegraphics[width=\linewidth]{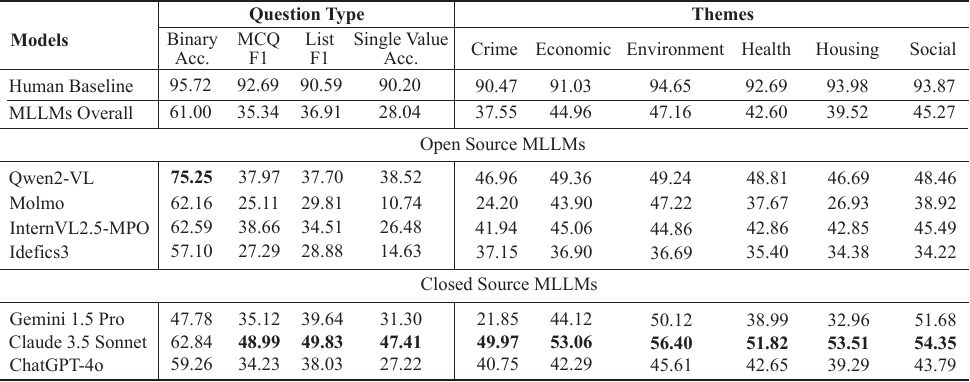}
\caption{Performance of Humans and 7 MLLMs (overall and individual) across 4 Question Types and 6 Themes. Values represent average performance scores (in \%) computed for each question type and theme. Bolded values indicate the best-performing model for each experimental condition.}
\label{tab:overall-performance-map-type-theme}
\vspace{-4mm}
\end{figure}

\emph{Task Type:} As shown in Fig~\ref{tab:overall-performance-task-type-question-type}, Claude 3.5 Sonnet emerged as the best-performing model in four out of six tasks, notably excelling in Regional Comparisons with an accuracy of 65.84\%. Qwen2-VL performed best in the Determine Range and Retrieve Value tasks, achieving particularly strong performance in Retrieve Value (61.89\%). In contrast, Molmo and Idefics3 consistently showed the weakest performance, with Idefics3 recording the lowest overall accuracy (15.56\%) in Determine Range. Overall, MLLMs performed best in Regional Comparisons, followed by Spatial Extremes, Pairwise Point Comparisons, Retrieve Value, Spatial Clusters, and Determine Range. Interestingly, despite its relative simplicity—requiring only straightforward extraction of visual encoding information for a single state—most MLLMs encountered significant difficulties with the Retrieve Value task. Conversely, Regional Comparisons, which demand comprehension of broader spatial patterns and zoning information, appeared comparatively easier for MLLMs. 

\emph{Map Type:} Across all three map types, Claude 3.5 Sonnet consistently emerged as the highest-performing model, see details in Fig~\ref{tab:overall-performance-map-type-theme}. In contrast, Molmo and Idefics3 consistently ranked among the weakest performers, with Idefics3 notably struggling on Cartograms, achieving an accuracy of only 33.63\%. 
Overall, MLLMs generally performed best on Choropleth maps, followed by Cartograms and Proportional Symbol maps. Notable exceptions include Claude 3.5 Sonnet and Molmo, which exhibited slightly higher accuracy on Cartograms, possibly due to enhanced color differentiation capabilities facilitated by uniform spatial unit sizes. Another exception was Gemini 1.5 Pro, which showed its highest accuracy on Proportional Symbol maps, highlighting its relatively stronger capability in differentiating symbol sizes as opposed to color variations. 

\emph{Theme}: Similar to the results by map type, Claude 3.5 Sonnet consistently emerged as the top-performing model across all themes, while Molmo and Idefics3 generally ranked among the lowest performers (Fig~\ref{tab:overall-performance-map-type-theme}). Notably, Gemini 1.5 Pro exhibited the weakest performance within the Crime theme, with an accuracy of 21.85\%, marking the lowest result among all evaluated models and themes. Further analysis revealed that MLLMs achieved the highest accuracy on maps depicting environmental data, followed by social, economic, health, housing, and crime-related maps. Model performance varied substantially across these themes, ranging from 47.16\% (environment) to 37.55\% (crime). This significant variation indicates a possible bias toward certain thematic topics, potentially due to models leveraging internal, topic-specific knowledge.

\emph{Question Type:} In general, MLLMs performed best on binary questions, followed by list, MCQ, and single-value question types. An intriguing observation is that models performed slightly better on list questions than MCQs (36.91\% vs. 35.34\%), despite MCQs typically being considered easier by human standards, as demonstrated by our Human Baseline results. This indicates that while models can identify multiple relevant elements within a list, they may face difficulty selecting the single most appropriate option from multiple-choice scenarios. At the model level, Qwen2-VL notably excelled in binary questions, achieving an impressive accuracy of 75.25\% (Fig~\ref{tab:overall-performance-map-type-theme}). For all other question types—list, MCQ, and single-value—Claude 3.5 Sonnet consistently delivered the best performance, reaffirming its position as the most robust model overall.

\begin{figure}[t]
\centering	
\includegraphics[width=\linewidth]{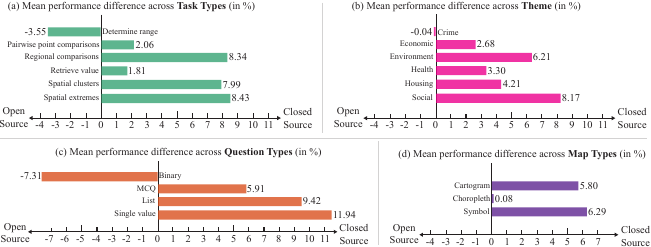}
\caption{Mean performance differences (in \%) between open- and closed-source models across four experimental variables: (a) Task Type, (b) Theme, (c) Question Type, and (d) Map Type. Bars extending to the right of zero indicate better performance by closed-source models, while bars to the left indicate better performance by open-source models.}
\label{fig:open-closed-models-compare}
\vspace{-4mm}
\end{figure}

\noindent \textbf{Comparing closed-sourced and open-sourced MLLMs} - Comparing differences in mean performance scores across experimental variables provided a detailed view of how closed-source models fare against open-source ones. As shown in Fig~\ref{fig:open-closed-models-compare}(a), closed-source models generally performed better, though the average gap was modest at 4.18\%. Notably, open-source models outperformed closed-source ones in the Determine Range task, largely due to ChatGPT-4o’s poor performance (16.30\%), the second-lowest overall (Fig~\ref{tab:overall-performance-task-type-question-type}). For map types (Fig~\ref{fig:open-closed-models-compare}(d)), the largest gap (6.29\%) was in Proportional Symbol maps, indicating that closed-source models are better at interpreting symbol size as a visual encoding. The smallest gap (0.08\%) appeared in Choropleth maps, suggesting similar proficiency, likely due to their frequent inclusion in open-source training data.
Furthermore, open-source models outperformed closed-source models on binary questions (Fig~\ref{fig:open-closed-models-compare}(c)), mainly due to Qwen2-VL’s strong performance and Gemini 1.5 Pro’s weaker showing (47.78\%). Thematic differences were also notable as Fig~\ref{fig:open-closed-models-compare}(b) shows: larger gaps appeared in social and environmental themes, while performance was similar in crime. This variation may reflect closed-source models' broader internal knowledge, aiding the interpretation of certain topics.
On average, the performance gap across all variables remained modest at around 4.4\%. These results suggest that open- and closed-source models are becoming increasingly comparable in interpreting thematic maps. While closed-source models hold a slight edge, the rapid progress of open-source models—especially Qwen2-VL—highlights the strong potential for collaborative advances in geospatial reasoning and map-based visual question answering.

\noindent \textbf{Comparing MLLMs with Human Baseline} -
While open and closed source models demonstrate somewhat comparable performance, our results show that MLLMs performed considerably worse than the human baseline, with humans outperforming MLLMs by an average margin of 50.35\% across all four experimental variables. Upon analyzing specific task types, the largest performance gap (68.79\%) appeared in the Determine Range task. Human readers executed this task effectively, whereas MLLMs struggled to interpret visual encodings within localized regional contexts. 
With respect to map types, the greatest performance difference was observed in Proportional Symbol Maps (52.48\%) compared to Cartogram and Choropleth maps. 
Overall, this comparative analysis demonstrates that MLLMs are currently well below human levels of performance for MapVQA tasks.

\section{Map Design Variations}

We were also interested in determining how MLLM's robustness (i.e., performance degradation relative to a baseline map design) and sensitivity (i.e., performance fluctuations across different design changes) are impacted by variations in map design. 

\begin{figure}[t]
\centering	
\includegraphics[width=0.9\linewidth]{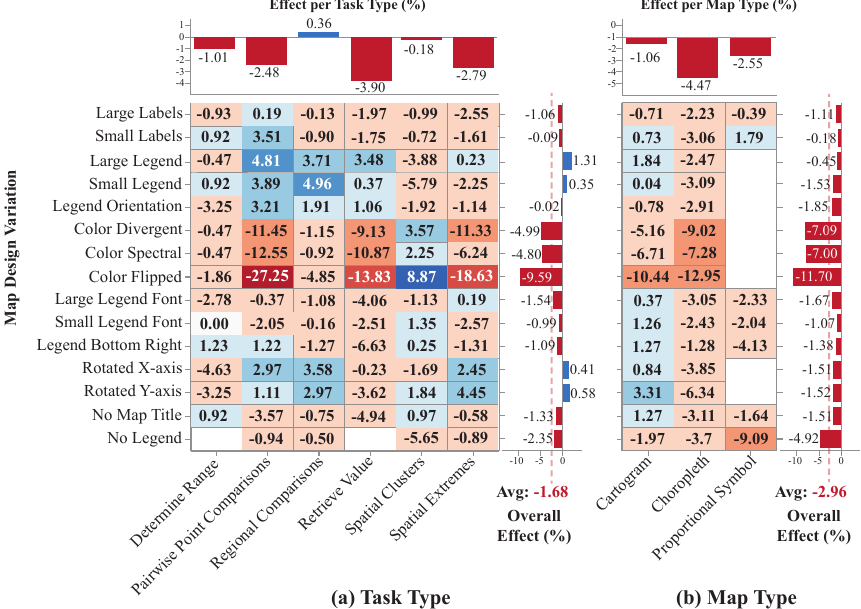}
\caption{Difference in performance of Qwen2-VL relative to the baseline across 15 map design variations, broken down by (a) Task Type and (b) Map Type. Each cell represents the percentage change in accuracy for a specific task or map type under a given variation.}
\label{fig:performance-Qwen}
\vspace{-4mm}
\end{figure}

\begin{figure}[t]
\centering	
\includegraphics[width=0.9\linewidth]{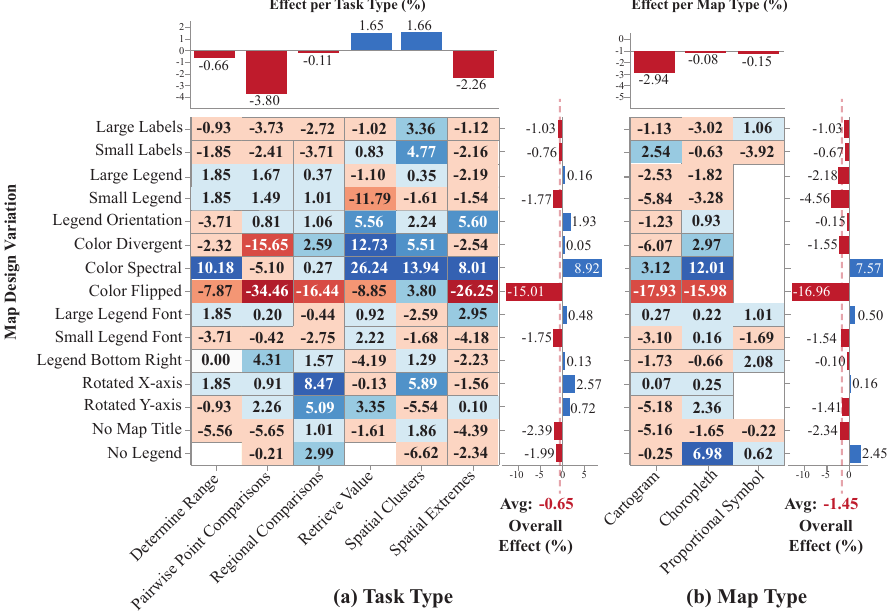}
\caption{Difference in performance of Claude 3.5 Sonnet relative to the baseline across 15 map design variations, divided by (a) Task Type and (b) Map Type. Each cell represents the percentage change in accuracy for a specific task or map type under a given variation.}
\label{fig:performance-Claude}
\vspace{-4mm}
\end{figure}

\subsection{Variations Tested}

We selected 15 map design variations targeting visual elements essential to map-reading tasks, spanning five categories: label size (Large vs. Small), legend properties (size, font, orientation, placement), color schemes (divergent, spectral, flipped), map orientation (180° X and Y-axis rotations), and removal of key elements (title, legend). Divergent and spectral schemes from ColorBrewer~\citep{harrower2003colorbrewer} were chosen for their effectiveness in class separation~\citep{slocum2022thematic}, enabling evaluation of model behavior under non-default color encodings. The flipped scheme (lighter shades = higher classes) tested robustness against counterintuitive mappings. Title and legend removals assessed the models' ability to rely solely on visual encoding without supplementary textual context, while modifications to label and legend properties examined sensitivity to minor visual changes. Axis rotations were introduced to evaluate potential “north-up” bias and the effect of orientation on map-reading performance. Some variations were not uniformly applicable across all map types due to differences in visual encoding structures (e.g., color schema changes did not apply to proportional symbol maps). Additionally, for maps without titles or legends, minor prompt edits (e.g., “darker shades indicate higher class values”) ensured fair evaluation. Further information on exceptions and an illustrative example is provided in Appendix~\ref{A3_map_design_variations}. 

To manage computational requirements, we systematically applied these variations to a representative subset of 36 maps, selected by randomly choosing two topics from each of the six themes across all three map types in the MapIQ dataset. This resulted in 540 uniquely varied maps (36 maps × 15 variations), accompanied by 684 QA pairs, with all maps provided in the supplementary materials. Baseline performance was first established using the unmodified maps, after which the test dataset was evaluated using Claude 3.5 Sonnet and Qwen2-VL—the top-performing closed- and open-source models from our benchmarking experiment.
\subsection{Results and Analysis}

We computed the performance difference for each variation relative to the baseline accuracy, separately by Task Type and Map Type. Overall, Claude 3.5 Sonnet exhibited greater robustness to design perturbations than its open-source counterpart, Qwen2-VL. As shown in Fig~\ref{fig:performance-Qwen}(a), Qwen2-VL experienced an average performance degradation of -1.68\% across task types, with 11 out of 15 variations negatively impacting performance. At the map type level (Fig~\ref{fig:performance-Qwen}(b)), the model showed an even larger average decline of -2.96\%, where all 15 variations reduced accuracy. These results suggest a general vulnerability of Qwen2-VL to visual design changes.

Further analysis revealed that variations related to color schemes produced the most pronounced negative effects on Qwen2-VL across both task and map types—particularly the Color Flipped variation, which resulted in the steepest decline in performance. At the map level (Fig~\ref{fig:performance-Qwen}(b)), Choropleth maps saw the greatest performance drop, possibly due to their over-representation in the model's training data. For Proportional Symbol maps, the most severe drop occurred under the No Legend condition (-9.09\%), highlighting the model’s reliance on legends for interpreting symbol sizes. Conversely, the Small Labels variation led to a modest improvement (+1.79\%), likely due to reduced visual clutter. Notably, altering legend font size also caused a performance drop across task and map types, suggesting \textbf{difficulty in interpreting complex or non-standard legends}. These patterns collectively point to \textbf{insufficient visual grounding} and \textbf{heightened sensitivity to design variations} in Qwen2-VL.

In comparison, Claude 3.5 Sonnet exhibited a smaller average degradation of -0.65\% across task types (Fig~\ref {fig:performance-Claude}(a)), with only 7 out of 15 variations negatively impacting performance. Fig~\ref {fig:performance-Claude}(a) further indicates that tasks requiring comprehension of class hierarchies, such as Spatial Extremes and Pairwise Point Comparisons, were disproportionately affected. In contrast, Spatial Clusters—which depend more on spatial proximity than value ordering—showed improved performance. Across map types (Fig~\ref {fig:performance-Claude}(b)), Claude showed an average drop of -1.45\%, again lower than Qwen2-VL. As with Qwen2-VL, the Color Flipped variation caused the most significant performance decline. This drop exceeded that of
Qwen2-VL, indicating that \textbf{Claude may be more sensitive to disruptions in sequential
color schemes.} However, Claude exhibited a strong performance boost under the Color Spectral scheme, suggesting a \textbf{superior ability to interpret color encodings using distinct hues.} Interestingly, under the No Map Legend condition, Claude’s performance in Choropleth maps improved (+6.98\%), possibly indicating better reliance on visual encoding when legend information is absent. Claude also showed a sharper decline than Qwen2-VL under the No Title condition, perhaps reflecting a \textbf{greater dependence on internal contextual knowledge}—potentially due to its larger and more diverse training corpus.

\section{Limitations and Future Work}

While our study provides a benchmark for evaluating the Map-VQA ability of an MLLM, there are several key limitations. We exclusively employed zero-shot prompting to evaluate MLLM performance, without exploring alternative strategies such as role-play prompting or visual prompting. The geographic scope was limited to U.S. state-level resolution, restricting our ability to examine how spatial granularity influences MLLM reasoning. Moreover, we did not evaluate model robustness against misleading or deceptive maps—a critical dimension for understanding visual comprehension with limited contextual knowledge. The study was also limited regarding map type diversity and visual complexity. We focused on three standard thematic map types—Choropleth, Cartogram, and Proportional Symbol maps, leaving out other important variants such as Isopleth maps and non-contiguous cartograms. Future research should expand on these aspects by incorporating a broader range of map types, exploring finer geographic resolutions (e.g., state-county or city–district level maps), and investigating high-level tasks such as automated caption generation and insight-based map summarization. 

\section{Conclusions}
In this study, we introduced MapIQ, a diverse and thematically rich benchmark dataset designed to evaluate the low-level map-reading capabilities of state-of-the-art MLLMs. We observed notable variations in model performance across map themes, suggesting that prior knowledge about the topic influences effectiveness, with models underperforming on less familiar themes. Among the models evaluated, Claude 3.5 Sonnet stood out for its superior accuracy and greater robustness to design variations compared to other MLLMs; however, MLLMs still lag behind human baselines. These findings highlight the need for more generalized and context-aware MLLMs—laying the groundwork for more trustworthy and interpretable visual reasoning systems.


\bibliography{colm2025_conference}

\begin{thebibliography}{51}
\providecommand{\natexlab}[1]{#1}
\providecommand{\url}[1]{\texttt{#1}}
\expandafter\ifx\csname urlstyle\endcsname\relax
  \providecommand{\doi}[1]{doi: #1}\else
  \providecommand{\doi}{doi: \begingroup \urlstyle{rm}\Url}\fi

\bibitem[Amar et~al.(2005)Amar, Eagan, and Stasko]{amar2005low}
Robert Amar, James Eagan, and John Stasko.
\newblock Low-level components of analytic activity in information visualization.
\newblock In \emph{IEEE Symposium on Information Visualization, 2005. INFOVIS 2005.}, pp.\  111--117. IEEE, 2005.

\bibitem[{Anthropic}(2024)]{anthropic_claude_2024}
{Anthropic}.
\newblock Claude 3.5 sonnet.
\newblock \url{https://www.anthropic.com/news/claude-3-5-sonnet}, 2024.

\bibitem[Bendeck \& Stasko(2024)Bendeck and Stasko]{bendeck2024empirical}
Alexander Bendeck and John Stasko.
\newblock An empirical evaluation of the gpt-4 multimodal language model on visualization literacy tasks.
\newblock \emph{IEEE Transactions on Visualization and Computer Graphics}, 2024.

\bibitem[Brehmer \& Munzner(2013)Brehmer and Munzner]{brehmer2013multi}
Matthew Brehmer and Tamara Munzner.
\newblock A multi-level typology of abstract visualization tasks.
\newblock \emph{IEEE transactions on visualization and computer graphics}, 19\penalty0 (12):\penalty0 2376--2385, 2013.

\bibitem[Brewer(2016)]{brewer2016designing}
Cynthia Brewer.
\newblock \emph{Designing Better Maps: A Guide for GIS Users, 2nd Edition}.
\newblock ESRI press, 2016.

\bibitem[{Centers for Disease Control and Prevention}(2024)]{cdc_data}
{Centers for Disease Control and Prevention}.
\newblock Cdc public health data, 2024.
\newblock URL \url{https://www.cdc.gov/places/index.html}.

\bibitem[Chang et~al.(2022)Chang, Palzer, Li, Fosler-Lussier, and Xiao]{chang2022mapqa}
Shuaichen Chang, David Palzer, Jialin Li, Eric Fosler-Lussier, and Ningchuan Xiao.
\newblock Mapqa: A dataset for question answering on choropleth maps.
\newblock \emph{arXiv preprint arXiv:2211.08545}, 2022.

\bibitem[Chen et~al.(2024)Chen, Wu, Wang, Su, Chen, Xing, Zhong, Zhang, Zhu, Lu, et~al.]{chen2024internvl}
Zhe Chen, Jiannan Wu, Wenhai Wang, Weijie Su, Guo Chen, Sen Xing, Muyan Zhong, Qinglong Zhang, Xizhou Zhu, Lewei Lu, et~al.
\newblock Internvl: Scaling up vision foundation models and aligning for generic visual-linguistic tasks.
\newblock In \emph{Proceedings of the IEEE/CVF Conference on Computer Vision and Pattern Recognition}, pp.\  24185--24198, 2024.

\bibitem[Deitke et~al.(2024)Deitke, Clark, Lee, Tripathi, Yang, Park, Salehi, Muennighoff, Lo, Soldaini, et~al.]{deitke2024molmo}
Matt Deitke, Christopher Clark, Sangho Lee, Rohun Tripathi, Yue Yang, Jae~Sung Park, Mohammadreza Salehi, Niklas Muennighoff, Kyle Lo, Luca Soldaini, et~al.
\newblock Molmo and pixmo: Open weights and open data for state-of-the-art multimodal models.
\newblock \emph{arXiv preprint arXiv:2409.17146}, 2024.

\bibitem[Dent(1999)]{dent1999cartography}
Borden~D Dent.
\newblock \emph{Cartography: Thematic Map Design}.
\newblock Ingram, 1999.

\bibitem[Derczynski(2016)]{derczynski2016complementarity}
Leon Derczynski.
\newblock Complementarity, f-score, and nlp evaluation.
\newblock In \emph{Proceedings of the Tenth International Conference on Language Resources and Evaluation (LREC'16)}, pp.\  261--266, 2016.

\bibitem[Fan et~al.(2024)Fan, Lei, Mancenido, Maceachren, and Maciejewski]{fan2024understanding}
Arlen Fan, Fan Lei, Michelle Mancenido, Alan~M Maceachren, and Ross Maciejewski.
\newblock Understanding reader takeaways in thematic maps under varying text, detail, and spatial autocorrelation.
\newblock In \emph{Proceedings of the 2024 CHI Conference on Human Factors in Computing Systems}, pp.\  1--17, 2024.

\bibitem[{Federal Aviation Administration}(2024)]{stateAbbrev}
{Federal Aviation Administration}.
\newblock Two-letter state and territory abbreviations, 2024.
\newblock URL \url{https://www.faa.gov/air_traffic/publications/atpubs/cnt_html/appendix_a.html}.

\bibitem[{Federal Bureau of Investigation}(2024)]{fbi_crime_data_explorer}
{Federal Bureau of Investigation}.
\newblock Crime data explorer, 2024.
\newblock URL \url{https://cde.ucr.cjis.gov/}.

\bibitem[Gemini(2024)]{team2024gemini}
Google Gemini.
\newblock Gemini 1.5: Unlocking multimodal understanding across millions of tokens of context.
\newblock \emph{arXiv preprint arXiv:2403.05530}, 2024.

\bibitem[Harrower \& Brewer(2003)Harrower and Brewer]{harrower2003colorbrewer}
Mark Harrower and Cynthia~A Brewer.
\newblock Colorbrewer. org: an online tool for selecting colour schemes for maps.
\newblock \emph{The Cartographic Journal}, 40\penalty0 (1):\penalty0 27--37, 2003.

\bibitem[Herrmann \& Pickle(1996)Herrmann and Pickle]{herrmann1996cognitive}
Douglas Herrmann and Linda~Williams Pickle.
\newblock A cognitive subtask model of statistical map reading.
\newblock \emph{Visual Cognition}, 3\penalty0 (2):\penalty0 165--190, 1996.

\bibitem[Hochmair et~al.(2024)Hochmair, Juh{\'a}sz, and Kemp]{hochmair2024correctness}
Hartwig~H Hochmair, Levente Juh{\'a}sz, and Takoda Kemp.
\newblock Correctness comparison of chatgpt-4, gemini, claude-3, and copilot for spatial tasks.
\newblock \emph{Transactions in GIS}, 28\penalty0 (7):\penalty0 2219--2231, 2024.

\bibitem[{Hugging Face}(2024)]{huggingface_2024}
{Hugging Face}.
\newblock Hugging face: The ai community building the future.
\newblock \url{https://huggingface.co/}, 2024.

\bibitem[Jenks \& Caspall(1971)Jenks and Caspall]{jenks1971error}
George~F Jenks and Fred~C Caspall.
\newblock Error on choroplethic maps: definition, measurement, reduction.
\newblock \emph{Annals of the Association of American Geographers}, 61\penalty0 (2):\penalty0 217--244, 1971.

\bibitem[Kafle et~al.(2018)Kafle, Price, Cohen, and Kanan]{kafle2018dvqa}
Kushal Kafle, Brian Price, Scott Cohen, and Christopher Kanan.
\newblock Dvqa: Understanding data visualizations via question answering.
\newblock In \emph{Proceedings of the IEEE conference on computer vision and pattern recognition}, pp.\  5648--5656, 2018.

\bibitem[Laurençon et~al.(2024)Laurençon, Marafioti, Sanh, and Tronchon]{laurençon2024building}
Hugo Laurençon, Andrés Marafioti, Victor Sanh, and Léo Tronchon.
\newblock Building and better understanding vision-language models: insights and future directions., 2024.

\bibitem[Lee et~al.(2016)Lee, Kim, and Kwon]{lee2016vlat}
Sukwon Lee, Sung-Hee Kim, and Bum~Chul Kwon.
\newblock Vlat: Development of a visualization literacy assessment test.
\newblock \emph{IEEE Transactions on Visualization and Computer Graphics}, 23\penalty0 (1):\penalty0 551--560, 2016.

\bibitem[Lu et~al.(2022)Lu, Mishra, Xia, Qiu, Chang, Zhu, Tafjord, Clark, and Kalyan]{lu2022learn}
Pan Lu, Swaroop Mishra, Tanglin Xia, Liang Qiu, Kai-Wei Chang, Song-Chun Zhu, Oyvind Tafjord, Peter Clark, and Ashwin Kalyan.
\newblock Learn to explain: Multimodal reasoning via thought chains for science question answering.
\newblock \emph{Advances in Neural Information Processing Systems}, 35:\penalty0 2507--2521, 2022.

\bibitem[MacEachren(2004)]{maceachren2004maps}
Alan~M MacEachren.
\newblock \emph{How maps work: representation, visualization, and design}.
\newblock Guilford Press, 2004.

\bibitem[Masry et~al.(2022)Masry, Long, Tan, Joty, and Hoque]{masry2022chartqa}
Ahmed Masry, Do~Xuan Long, Jia~Qing Tan, Shafiq Joty, and Enamul Hoque.
\newblock Chartqa: A benchmark for question answering about charts with visual and logical reasoning.
\newblock \emph{arXiv preprint arXiv:2203.10244}, 2022.

\bibitem[Mukhopadhyay et~al.(2024{\natexlab{a}})Mukhopadhyay, Qidwai, Garimella, Ramu, Gupta, and Roth]{mukhopadhyay2024unraveling}
Srija Mukhopadhyay, Adnan Qidwai, Aparna Garimella, Pritika Ramu, Vivek Gupta, and Dan Roth.
\newblock Unraveling the truth: Do vlms really understand charts? a deep dive into consistency and robustness.
\newblock In \emph{Findings of the Association for Computational Linguistics: EMNLP 2024}, pp.\  16696--16717, 2024{\natexlab{a}}.

\bibitem[Mukhopadhyay et~al.(2024{\natexlab{b}})Mukhopadhyay, Rajgaria, Khatiwada, Gupta, and Roth]{mukhopadhyay2024mapwise}
Srija Mukhopadhyay, Abhishek Rajgaria, Prerana Khatiwada, Vivek Gupta, and Dan Roth.
\newblock Mapwise: Evaluating vision-language models for advanced map queries.
\newblock \emph{arXiv preprint arXiv:2409.00255}, 2024{\natexlab{b}}.

\bibitem[Munzner(2014)]{munzner2014visualization}
Tamara Munzner.
\newblock \emph{Visualization analysis and design}.
\newblock CRC press, 2014.

\bibitem[{National Weather Service}(2024)]{weather_gov}
{National Weather Service}.
\newblock Weather.gov - national weather data, 2024.
\newblock URL \url{https://www.weather.gov/}.

\bibitem[{OpenAI}(2024)]{openai_chatgpt4o_2024}
{OpenAI}.
\newblock Hello gpt-4o.
\newblock \url{https://openai.com/index/hello-gpt-4o/}, 2024.

\bibitem[Pandey \& Ottley(2023)Pandey and Ottley]{pandey2023mini}
Saugat Pandey and Alvitta Ottley.
\newblock Mini-vlat: A short and effective measure of visualization literacy.
\newblock In \emph{Computer Graphics Forum}, volume~42, pp.\  1--11. Wiley Online Library, 2023.

\bibitem[Radford et~al.(2019)Radford, Wu, Child, Luan, Amodei, Sutskever, et~al.]{radford2019language}
Alec Radford, Jeffrey Wu, Rewon Child, David Luan, Dario Amodei, Ilya Sutskever, et~al.
\newblock Language models are unsupervised multitask learners.
\newblock \emph{OpenAI blog}, 1\penalty0 (8):\penalty0 9, 2019.

\bibitem[Roberts et~al.(2024)Roberts, L{\"u}ddecke, Sheikh, Han, and Albanie]{roberts2024charting}
Jonathan Roberts, Timo L{\"u}ddecke, Rehan Sheikh, Kai Han, and Samuel Albanie.
\newblock Charting new territories: Exploring the geographic and geospatial capabilities of multimodal llms.
\newblock In \emph{Proceedings of the IEEE/CVF Conference on Computer Vision and Pattern Recognition}, pp.\  554--563, 2024.

\bibitem[Rogerson(2019)]{rogerson2019statistical}
Peter~A Rogerson.
\newblock \emph{Statistical methods for geography: a student's guide}.
\newblock SAGE Publications Ltd, 2019.

\bibitem[Slocum et~al.(2022)Slocum, McMaster, Kessler, and Howard]{slocum2022thematic}
Terry~A Slocum, Robert~B McMaster, Fritz~C Kessler, and Hugh~H Howard.
\newblock \emph{Thematic Cartography and Geovisualization}.
\newblock CRC Press, 2022.

\bibitem[Snyder(1982)]{snyder1982map}
John~Parr Snyder.
\newblock Map projections used by the {U.S.} {G}eological {S}urvey.
\newblock Technical report, US Government Printing Office, 1982.

\bibitem[Thomson et~al.(2005)Thomson, Hetzler, MacEachren, Gahegan, and Pavel]{thomson2005typology}
Judi Thomson, Elizabeth Hetzler, Alan MacEachren, Mark Gahegan, and Misha Pavel.
\newblock A typology for visualizing uncertainty.
\newblock In \emph{Visualization and Data Analysis}, volume 5669, pp.\  146--157, 2005.

\bibitem[{U.S. Census Bureau}(2022)]{uscb_econ_geo_levels}
{U.S. Census Bureau}.
\newblock Geographic levels, 2022.
\newblock URL \url{https://www.census.gov/programs-surveys/economic-census/guidance-geographies/levels.html}.
\newblock Accessed: 2025-03-22.

\bibitem[{U.S. Census Bureau}(2024{\natexlab{a}})]{usZoning}
{U.S. Census Bureau}.
\newblock Census regions and divisions of the united states, 2024{\natexlab{a}}.
\newblock URL \url{https://www2.census.gov/geo/pdfs/maps-data/maps/reference/us_regdiv.pdf}.

\bibitem[{U.S. Census Bureau}(2024{\natexlab{b}})]{us_census_acs}
{U.S. Census Bureau}.
\newblock American community survey (acs), 2024{\natexlab{b}}.
\newblock URL \url{https://www.census.gov/programs-surveys/acs}.

\bibitem[{U.S. Environmental Protection Agency}(2024)]{epa_data}
{U.S. Environmental Protection Agency}.
\newblock Epa environmental data, 2024.
\newblock URL \url{https://www.epa.gov/aqs}.

\bibitem[Van~Rijsbergen(1974)]{van1974foundation}
Cornelis~Joost Van~Rijsbergen.
\newblock Foundation of evaluation.
\newblock \emph{Journal of documentation}, 30\penalty0 (4):\penalty0 365--373, 1974.

\bibitem[Wang et~al.(2024)Wang, Bai, Tan, Wang, Fan, Bai, Chen, Liu, Wang, Ge, Fan, Dang, Du, Ren, Men, Liu, Zhou, Zhou, and Lin]{Qwen2VL}
Peng Wang, Shuai Bai, Sinan Tan, Shijie Wang, Zhihao Fan, Jinze Bai, Keqin Chen, Xuejing Liu, Jialin Wang, Wenbin Ge, Yang Fan, Kai Dang, Mengfei Du, Xuancheng Ren, Rui Men, Dayiheng Liu, Chang Zhou, Jingren Zhou, and Junyang Lin.
\newblock Qwen2-vl: Enhancing vision-language model's perception of the world at any resolution.
\newblock \emph{arXiv preprint arXiv:2409.12191}, 2024.

\bibitem[Wu et~al.(2024{\natexlab{a}})Wu, Yan, Shen, Wang, Tang, and Luo]{wu2024chartinsights}
Yifan Wu, Lutao Yan, Leixian Shen, Yunhai Wang, Nan Tang, and Yuyu Luo.
\newblock Chartinsights: Evaluating multimodal large language models for low-level chart question answering.
\newblock \emph{arXiv preprint arXiv:2405.07001}, 2024{\natexlab{a}}.

\bibitem[Wu et~al.(2024{\natexlab{b}})Wu, Chen, Pan, Liu, Liu, Dai, Gao, Ma, Wu, Wang, Xie, Wu, Hu, Wang, Sun, Li, Piao, Guan, Liu, Xie, You, Dong, Yu, Zhang, Zhao, Wang, and Ruan]{wu2024deepseekvl2mixtureofexpertsvisionlanguagemodels}
Zhiyu Wu, Xiaokang Chen, Zizheng Pan, Xingchao Liu, Wen Liu, Damai Dai, Huazuo Gao, Yiyang Ma, Chengyue Wu, Bingxuan Wang, Zhenda Xie, Yu~Wu, Kai Hu, Jiawei Wang, Yaofeng Sun, Yukun Li, Yishi Piao, Kang Guan, Aixin Liu, Xin Xie, Yuxiang You, Kai Dong, Xingkai Yu, Haowei Zhang, Liang Zhao, Yisong Wang, and Chong Ruan.
\newblock Deepseek-vl2: Mixture-of-experts vision-language models for advanced multimodal understanding, 2024{\natexlab{b}}.
\newblock URL \url{https://arxiv.org/abs/2412.10302}.

\bibitem[Xing et~al.(2025)Xing, Sun, Xie, Chen, Huang, Wang, Li, Song, and Tu]{xing2025can}
Shuo Xing, Zezhou Sun, Shuangyu Xie, Kaiyuan Chen, Yanjia Huang, Yuping Wang, Jiachen Li, Dezhen Song, and Zhengzhong Tu.
\newblock Can large vision language models read maps like a human?
\newblock \emph{arXiv preprint arXiv:2503.14607}, 2025.

\bibitem[Xu \& Wall(2024)Xu and Wall]{xu2024exploring}
Zhongzheng Xu and Emily Wall.
\newblock Exploring the capability of llms in performing low-level visual analytic tasks on svg data visualizations.
\newblock In \emph{2024 IEEE Visualization and Visual Analytics (VIS)}, pp.\  126--130. IEEE, 2024.

\bibitem[Yao et~al.(2024)Yao, Yu, Zhang, Wang, Cui, Zhu, Cai, Li, Zhao, He, et~al.]{yao2024minicpm}
Yuan Yao, Tianyu Yu, Ao~Zhang, Chongyi Wang, Junbo Cui, Hongji Zhu, Tianchi Cai, Haoyu Li, Weilin Zhao, Zhihui He, et~al.
\newblock Minicpm-v: A gpt-4v level mllm on your phone.
\newblock \emph{arXiv preprint arXiv:2408.01800}, 2024.

\bibitem[Zhang et~al.(2023)Zhang, Zhang, Gu, Zhou, Lipka, Yang, and Sun]{zhang2023llavar}
Yanzhe Zhang, Ruiyi Zhang, Jiuxiang Gu, Yufan Zhou, Nedim Lipka, Diyi Yang, and Tong Sun.
\newblock Llavar: Enhanced visual instruction tuning for text-rich image understanding.
\newblock \emph{arXiv preprint arXiv:2306.17107}, 2023.

\bibitem[Zhou et~al.(2023)Zhou, Liu, Gu, Zou, Huang, Wu, Li, Chen, Zhou, Liu, et~al.]{zhou2023survey}
Hongjian Zhou, Fenglin Liu, Boyang Gu, Xinyu Zou, Jinfa Huang, Jinge Wu, Yiru Li, Sam~S Chen, Peilin Zhou, Junling Liu, et~al.
\newblock A survey of large language models in medicine: Progress, application, and challenge.
\newblock \emph{arXiv preprint arXiv:2311.05112}, 2023.

\end{thebibliography}
\bibliographystyle{colm2025_conference}

\appendix

\section{Appendix}
\subsection*{Supplementary Materials}
\label{supplementary_materials}
All supplementary materials are available at OSF: \url{https://osf.io/kp6j4/?view_only=fa2847a270094fd98512127bebd8de87}.

\subsection{More on MapIQ Dataset}
\subsubsection{Metadata Classified by Theme}
\label{A1.1_metadata_details}

The metadata for MapIQ was sourced from reputable sources, which informed the topical focus of the maps. Map topic is a fundamental element of thematic maps, as it helps readers contextualize the information being visualized and interpret the content more effectively~\citep{herrmann1996cognitive}. However, due to data quality constraints, the number of datasets per map theme in the full MapIQ dataset was not balanced. To address this in our analysis, the test dataset was sampled to ensure an equal number of question instances from each theme. Table~\ref{tab:theme-topics} summarizes the number of datasets associated with each metadata theme.

\begin{table}[h]
\setlength{\arrayrulewidth}{0.3mm}
\renewcommand{\arraystretch}{1.3}
\begin{tabular}{lcl}
\hline
Theme & \# Datasets & Example Topic \\ \hline
Economic & 46 & Percentage of Population in the Labor Force \\
Housing & 68 & Percentage of rental units with monthly rent under \$500 \\
Social & 49 & Percentage of cohabiting couple households \\
Health & 40 & Cognitive disability among adults (in percent) \\
Crime & 23 & Burglary Offenses per 100,000 People \\
Environment & 32 & Annual CO2 Emissions (Million Metric Tons) \\ \hline
\end{tabular}
\caption{Distribution of datasets by Theme, along with example map topics corresponding to each category.}
\label{tab:theme-topics}
\end{table}

\subsubsection{Task Type Definitions}
\label{A1.2_task_type_definitions}



MapIQ uses six distinct visual analytical tasks to evaluate the performance of MLLMs in map-reading contexts. Each task is clearly defined, emphasizing its geospatial scale (local or global), the complexity of the required visual analysis, and the specific cognitive skills involved.

\textbf{1. Retrieve Value: } This task involves identifying the attribute class of a specific state by referencing the map's legend. This fundamental map-reading task requires effectively linking legend classifications with the corresponding visual encodings on the map~\citep{lee2016vlat, maceachren2004maps}. The geospatial scale of this task is local, as it focuses on retrieving information about individual states rather than analyzing broader spatial patterns across the entire map.

\textbf{2. Pairwise Point Comparisons: } This task involves comparing the attribute classes of two specified states to determine whether one is greater or less than the other~\citep{lee2016vlat, amar2005low}. Pairwise Point Comparisons require interpreting and evaluating visual encodings for two distinct states simultaneously. Despite this added complexity, the geospatial scale remains local, as the analysis is limited solely to the two states in question and does not require an understanding of broader spatial patterns across the entire map.

\textbf{3. Spatial Extremes:} This task involves identifying states with either the highest or lowest attribute class values~\citep{lee2016vlat, munzner2014visualization}. Spatial Extremes can be assessed at both local and global geospatial scales~\citep{thomson2005typology}. The local version of the task requires finding extreme values within a specific map region (e.g., the West Zone), thereby limiting the analysis to a predefined regional subset. In contrast, the global version involves identifying extremes across the entire map, demanding a more comprehensive understanding of spatial patterns distributed throughout all states.

\textbf{4. Spatial Clusters:} This task involves identifying spatial clusters—groups of contiguous states sharing similar attribute values—which demands higher-level pattern recognition skills~\citep{lee2016vlat, maceachren2004maps}. Including this task required defining spatial contiguity, for which we adopted the queen adjacency rule, wherein states are considered neighbors if they share a common border or vertex~\citep{rogerson2019statistical}. Additionally, we focused specifically on clusters of states exhibiting extreme attribute values—clusters containing states with either consistently high or consistently low values—since identifying such "hot spots" or "cold spots" is among the most frequently executed spatial analytical tasks in practice~\citep{rogerson2019statistical}. We evaluated both local and global versions of this task. The local variant required identifying clusters within predefined map regions, while the global variant involved recognizing clusters spanning the entire map.

\textbf{5. Determine Range:} This task involves determining the range of attribute class values present within a specific map region~\citealp{munzner2014visualization, lee2016vlat}. This task is strictly local, as it confines the analysis exclusively to predefined subsets of the map. Executing this task demands careful identification of the minimum and maximum attribute class values among a selected group of states within the given region, effectively interpreting the visual encodings of multiple spatial units simultaneously. The inherent complexity arises from synthesizing information across the region rather than focusing on individual states, highlighting the task's reliance on precise legend interpretation and detailed visual analysis within localized contexts.

\textbf{6. Regional Comparisons:} This task involves comparing the overall attribute patterns between two distinct regions on the map to determine whether one region generally exhibits higher or lower attribute class values than the other~\citep{maceachren2004maps}. Unlike tasks that focus solely on individual states or single-region analyses, Regional Comparisons require synthesizing and evaluating broader spatial patterns across multiple states within each of the two selected regions. Among the tasks examined, this is the most open-ended, demanding strong visual pattern recognition and correlation skills to effectively discern general trends and differences between regions. This task is global in scale, involving reasoning about spatial configuration and distribution patterns across the map rather than isolated units~\citep{maceachren2004maps}.

\subsubsection{Question Templates}
\label{A1.3_question_templates_and_task_eligibility}

\begin{figure}[h]
\centering	
\includegraphics[width=\linewidth]{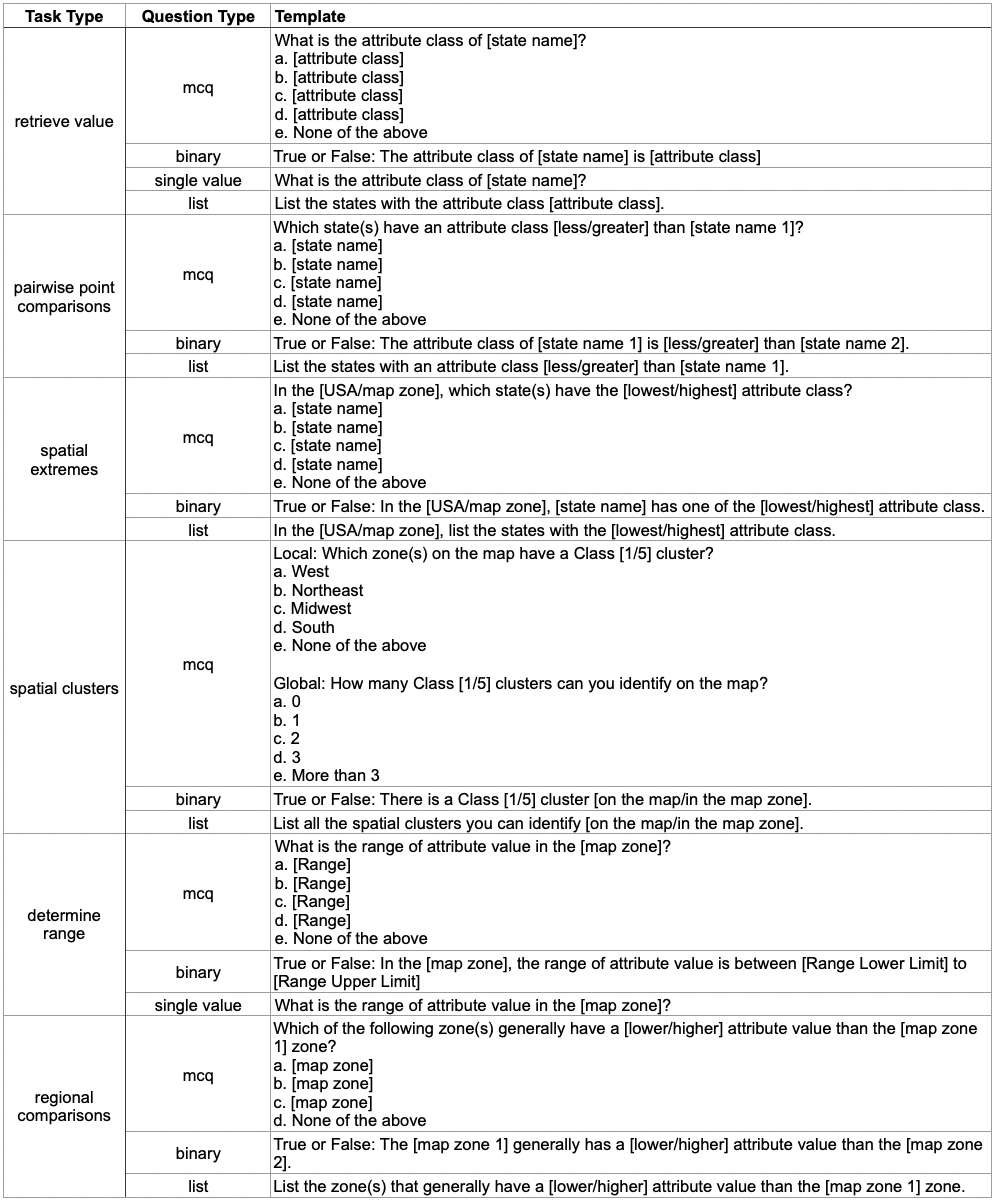}
\caption{Question templates for each Task Type, organized by corresponding Question Type.}
\label{fig:question-template}
\end{figure}

\begin{figure}[t]
\centering	
\includegraphics[width=0.8\linewidth]{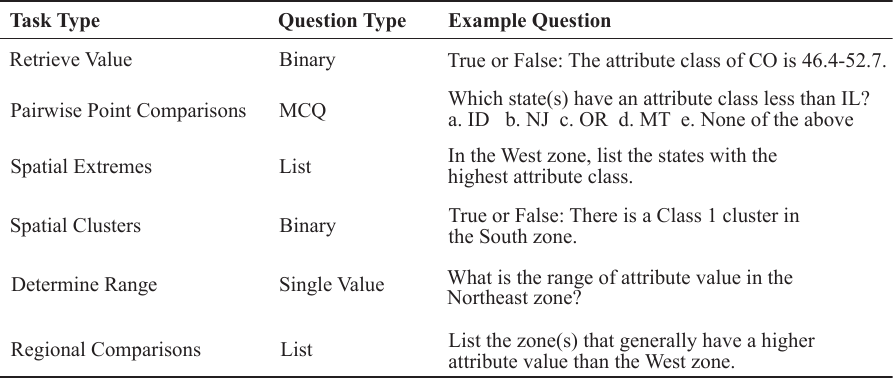}
\caption{Example questions across different task types and formats in the MapIQ dataset}
\label{tab:example-questions-table}
\end{figure}

Due to the large dataset size, questions for MapIQ could not be generated manually. Instead, we developed a set of manually crafted templates for each task and corresponding question type, as illustrated in Figure~\ref{fig:question-template}. A total of 19 unique question templates were created, with placeholders such as [attribute class], [state name], and [range] randomly populated during generation. Additionally, the placeholder [USA/map zone] was used to determine the spatial scale of tasks, allowing us to investigate both local and global spatial analysis. Example questions generated using these templates can be seen in Figure~\ref{tab:example-questions-table}. Furthermore, due to the varying nature of the tasks, not every question type was eligible for all task types; these exceptions are summarized in Table~\ref{tab:eligible-task-question-type}.

\begin{table}[]
\centering
\renewcommand{\arraystretch}{1.3}
\small
\begin{tabular}{ccccccc}
\hline
\multirow{2}{*}{\textbf{Task Type}} & \multicolumn{1}{c|}{\multirow{2}{*}{\textbf{Geospatial Scale}}} & \multicolumn{4}{c|}{\textbf{Question Type}} & \multirow{2}{*}{\textbf{Total}} \\ \cline{3-6}
 & \multicolumn{1}{c|}{} & Binary & MCQ & List & \multicolumn{1}{c|}{Single Value} &  \\ \hline
Retrieve Value & Local & \cmark\ & \cmark\ & \cmark\ & \cmark\ & 4 \\
Pairwise Point Comparisons & Local & \cmark\ & \cmark\ & \cmark\ & \xmark\ & 3 \\
Spatial Extremes & Local and Global & \cmark\ & \cmark\ & \cmark\ & \xmark\ & 3 \\
Spatial Clusters & Local and Global & \cmark\ & \cmark\ & \cmark\ & \xmark\ & 3 \\
Determine Range & Local & \cmark\ & \cmark\ & \xmark\ & \cmark\ & 3 \\
Regional Comparisons & Global & \cmark\ & \cmark\ & \cmark\ & \xmark\ & 3 \\ \hline
\end{tabular}
\caption{Eligibility of each task type for different question types, along with the associated geospatial scale (local or global). The "Total" column indicates the number of question types applicable to each task.}
\label{tab:eligible-task-question-type}
\end{table}


\subsubsection{Ground Truth Extraction and Validation}
\label{A1.4_ground_truth_extraction}

Ground truth answers were programmatically extracted from geospatial metadata (GeoJSON files) using Python scripts, without any visual map inspection. Given access to the original metadata and the objective, factual nature of our questions (e.g., "What is the attribute class of AL?"), programmatic extraction was both feasible and more accurate than manual annotation. Scripts were developed for each Task Type–Question Type combination to systematically extract the required information.

For example, for a multiple-choice question under the Retrieve Value task type, the script would parse the GeoJSON file to identify the map topic and target state (e.g., Alabama), look up the ground truth value (e.g., 11\% population mobility), match it to the correct option (e.g., "d. 10.0–11.2"), and return the corresponding ground truth answer (e.g., ['d']). We validated representative samples and refined the scripts iteratively based on consistent error patterns to ensure accuracy. The ground truth responses are provided in the supplementary materials (OSF: Folder Name – Ground Truth Responses).

\subsubsection{Comparing MapIQ with Existing Map-VQA Datasets}
\label{A1.5_comaring_with_existing_benchmarks}

\begin{figure}[t]
\centering	
\includegraphics[width=0.8\linewidth]{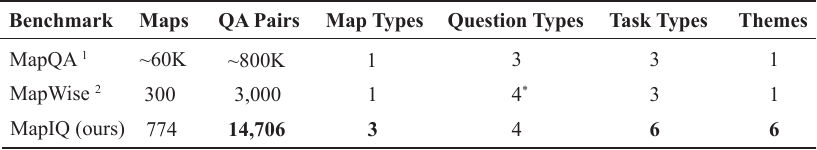}
\begin{flushleft}
{\footnotesize\textsuperscript{1} MapQA~\citep{chang2022mapqa}} 
{\footnotesize\textsuperscript{2} MapWise~\citep{mukhopadhyay2024mapwise}} 
{\footnotesize\textsuperscript{*}In MapWISE, the Single Word, Range, and Count question types all require open-text responses consisting of a single value. Therefore, they have been consolidated into the Single Value category, resulting in four distinct question types instead of the six originally reported.}
\end{flushleft}
\caption{Comparison of MapIQ with prior Map-VQA benchmarks.}
\label{tab:benchmark-comparison-table}
\end{figure}

Benchmark datasets focused on low-level map-reading tasks are still in their early stages, with limited coverage in terms of map diversity, task types, and topical themes. MapIQ advances this space by offering a more comprehensive benchmark that expands across three dimensions: a broader range of map types (including Choropleth, Cartogram, and Proportional Symbol maps), a diverse set of well-defined task types grounded in visual analytics, and thematically rich content drawn from real-world datasets. This breadth allows for a more nuanced evaluation of model capabilities across visual interpretation and contextual understanding. Figure~\ref{tab:benchmark-comparison-table} summarizes how MapIQ compares to existing benchmarks along these key dimensions.

\subsection{Benchmarking MLLMs}
\subsubsection{Prompt Details}
\label{A2.1_prompt_details}

\begin{figure}[h]
\centering	
\includegraphics[width=\linewidth]{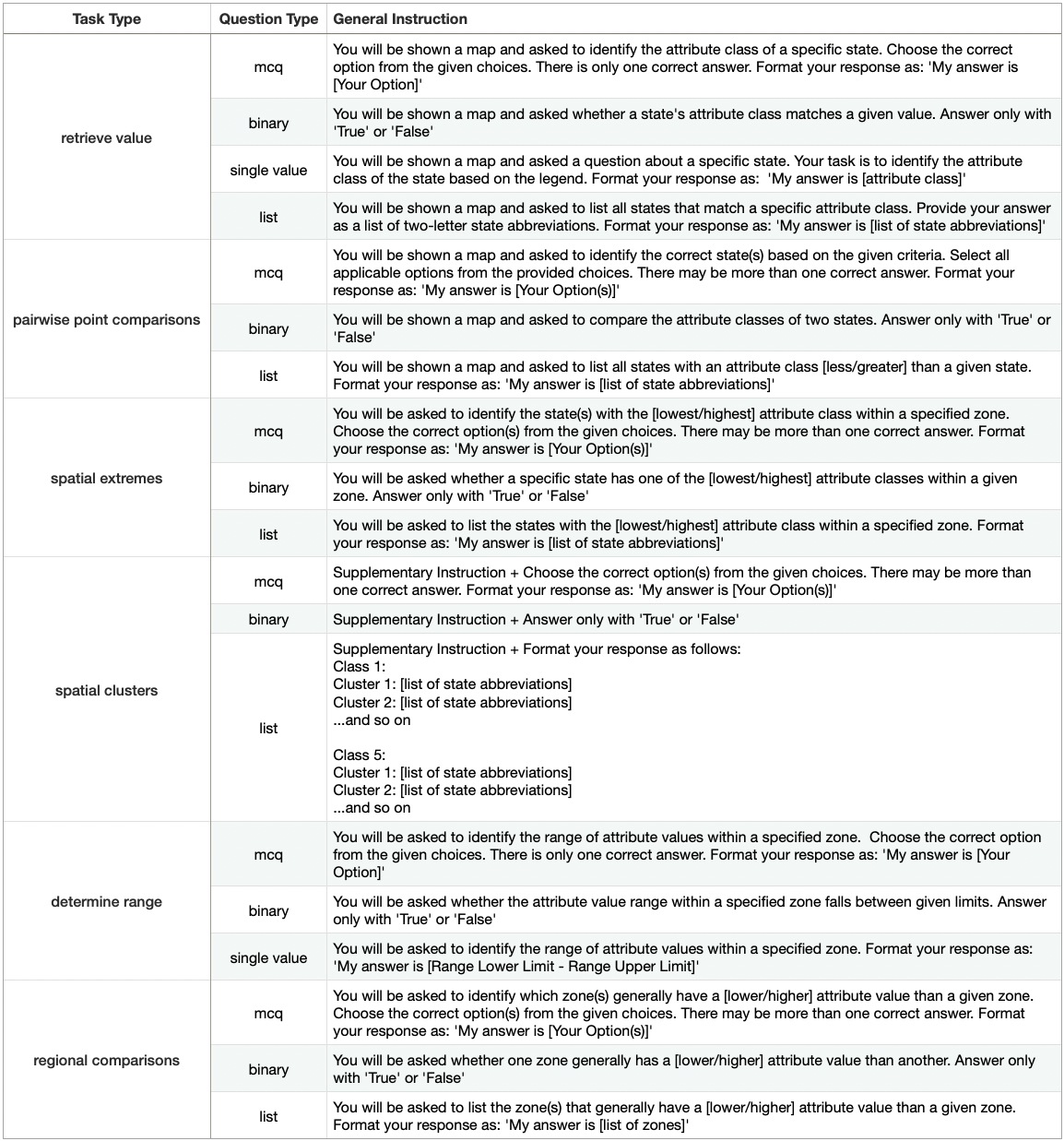}
\caption{General instructions provided as part of the prompts for each Task Type and Question Type, shared with MLLMs and human participants.}
\label{fig:prompts-example}
\end{figure}

Due to the specialized nature of our benchmark, we adopted a zero-shot prompting strategy tailored to our experimental setup. Each prompt followed a standardized format consisting of a zoning instruction (where applicable) followed by a general instruction. The general instructions used across tasks are illustrated in Figure~\ref{fig:prompts-example}. The zoning instruction introduced zone mapping for the contiguous U.S. states, which was stated as follows: “In the provided data, the states in the USA are classified into four zones: West, Midwest, Northeast, and South. Use this classification to answer a question based on a map.”

In addition, for the Spatial Clusters task, we incorporated supplementary information in the General Instruction, to define the concept of spatial cluster, stated as follows:
“A spatial cluster consists of geographically proximate locations sharing the same attribute class. You will be shown a map with data categorized into five classes and asked to identify spatial clusters. Only Class 1 (lowest) and Class 5 (highest) qualify as clusters, while Classes 2–4 do not. Clusters follow the Queen adjacency rule, meaning all states in the cluster must be connected without leaving the set. Each cluster must contain at least three states, with no upper limit, and multiple clusters can exist within the same class.”

\subsubsection{Test Set Sampling}
\label{A2.2_test_set_sampling}

For sampling the test set for the benchmarking experiment, we first identified all valid combinations of the experimental variables: map type, task type, question type, and theme. Since not all question types were compatible with every task type, we calculated the total number of valid combinations as follows:  
3 (map types) $\times$ 6 (task types) $\times$ 6 (themes) $\times$ 2 (question types applicable to all tasks—binary and MCQ) = 216 combinations;  
3 (map types) $\times$ 2 (task types: Retrieve Value and Determine Range) $\times$ 6 (themes) $\times$ 1 (question type: single-value) = 36 combinations;  
3 (map types) $\times$ 5 (all task types except Determine Range) $\times$ 6 (themes) $\times$ 1 (question type: list) = 90 combinations.  

\begin{figure}[t]
\centering	
\includegraphics[width=0.9\linewidth]{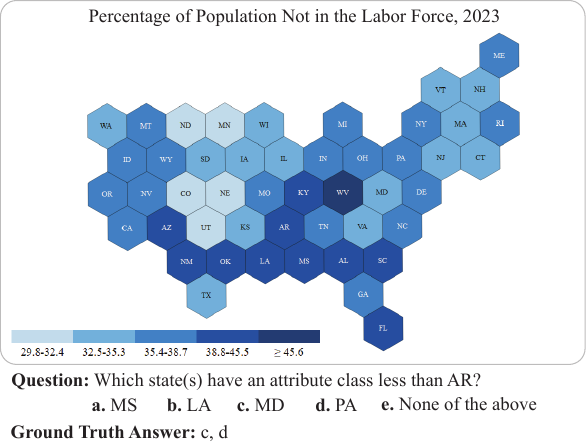}
\caption{Sample test dataset map with the following variable settings—Map Type: Cartogram, Task Type: Pairwise Point Comparisons, Question Type: MCQ, Theme: Economic. The map displays the question alongside multiple-choice options and the ground truth answer.}
\label{fig:test-map-example}
\end{figure}

This yielded a total of 342 unique combinations. We then selected 15 QA pairs per combination to ensure uniform representation, resulting in a balanced sample of 5130 QA pairs. For instance, the combination "map type: Cartogram, task type: Retrieve Value, theme: Economic, question type: MCQ" included exactly 15 QA pairs in the final dataset (see Fig~\ref{fig:test-map-example} for an example QA pair from this combination). To ensure topical diversity, map topics within each theme were randomly selected without replacement. While the test set maintained a perfect balance across map types and themes, minor imbalances remained across task and question types due to inherent compatibility limitations between certain tasks and question formats.

\subsubsection{MLLM Response Validation}
\label{A2.3_mllm_response_validation}

Manual postprocessing was performed across the entire test dataset for all model responses to ensure consistent and fair evaluation. Despite providing explicit formatting instructions (e.g., “Format your response as: 'My answer is [Your Option]'”), models frequently included extraneous content such as reasoning steps, explanations, or hallucinated information alongside their answers. Manual validation was essential to extract the actual answers, enforce consistent formatting, and prevent formatting variations from artificially penalizing model performance during evaluation.

For example, for a Determine Range question, Molmo returned: “To answer this question, I need to: [reasoning steps...] After examining the map and legend, Washington's circle falls into the second largest category, which corresponds to the 99.9–163 range. My answer is 99.9–163.” This was cleaned to [99.9, 163] for evaluation. Similarly, Idefics responded “No.” to a Binary question and was cleaned to FALSE for consistency. This process was repeated systematically across the entire test set to ensure accurate and reproducible evaluations. All raw model responses and their cleaned versions are provided in the Supplementary Materials (OSF: “MLLM Responses/Baseline”).

It is important to distinguish this validation process from the human baseline evaluation. Ground truth extraction and MLLM response validation were carried out by a single human, distinct from the two humans involved in human baseline evaluation.

\subsubsection{Sensitivity Comparison with Human Baseline}

\begin{figure}[t]
\centering	
\includegraphics[width=0.5\linewidth]{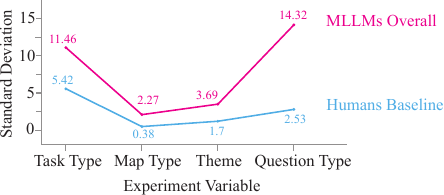}
\caption{The standard deviation of human baseline and MLLMs across experiment variables.}
\label{fig:human-model-std}
\end{figure}

We further analyzed the sensitivity of MLLM and human performance across the four primary experimental variables using the standard deviation of performance across variables. Figure~\ref{fig:human-model-std} shows that the standard deviation in MLLM performance consistently exceeded that of humans across all variables, indicating greater sensitivity to contextual changes. Notably, the highest variability was observed with respect to Task Type (11.46\%) and Question Type (14.32\%). While such variations were expected due to differences in task and question complexity, the significantly higher sensitivity of MLLMs suggests that these models are more affected by changes in difficulty than human readers. Additionally, whereas human performance remained relatively stable across different map types and themes, MLLMs showed more variability, suggesting potential biases toward specific map types and thematic content.


\subsection{Map Design Variations}

\subsubsection{Exceptions}
\label{A3_map_design_variations}

\begin{figure}[h]
\centering	
\includegraphics[width=\linewidth]{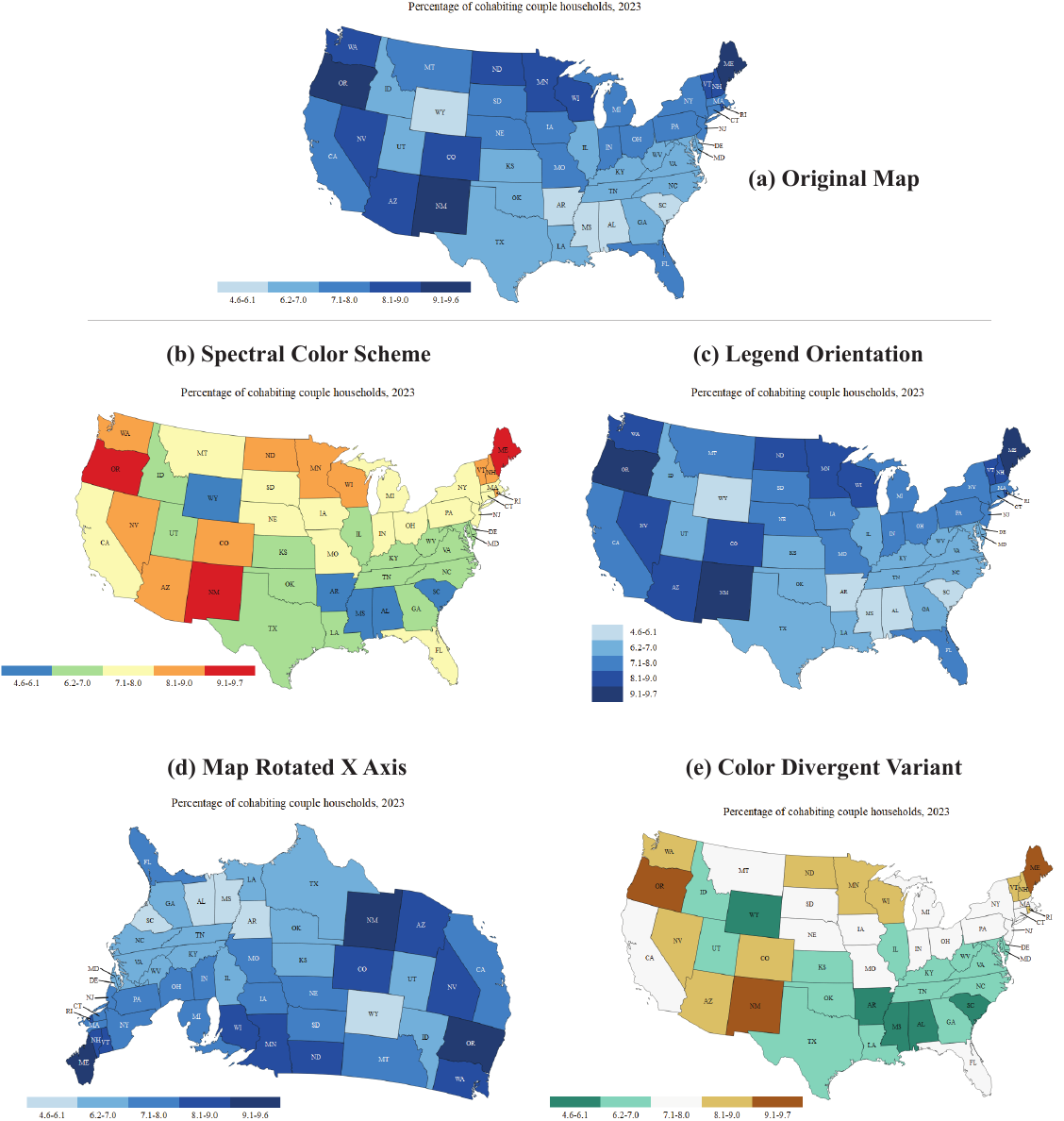}
\caption{Four examples of map design variations compared to the baseline design (a). (b) and (e) depict changes in color scheme, (c) illustrates a modification in legend orientation, and (d) shows a rotated map layout.}
\label{fig:map-variation}
\end{figure}

While we aimed to apply all 15 map design variations uniformly across all three map types and task types, differences in visual encoding made this infeasible. For instance, color scheme variations were not applied to Proportional Symbol maps, which rely on symbol size rather than color. Similarly, the Determine Range and Retrieve Value tasks require a legend, making the "No Legend" variation incompatible. To ensure fair evaluation in maps without titles or legends, minimal prompt edits (e.g., “darker shades indicate higher class values”) preserved essential contextual information. Figure~\ref{fig:map-variation} presents illustrative examples of the tested design variations.

\end{document}